\documentclass[a4paper]{article}

\usepackage[english]{babel}
\usepackage[utf8x]{inputenc}
\usepackage[T1]{fontenc}

\usepackage[a4paper,top=3cm,bottom=2cm,left=3cm,right=3cm,marginparwidth=1.75cm]{geometry}

\usepackage{amsmath}
\usepackage{graphicx}
\usepackage{authblk}
\usepackage[colorinlistoftodos]{todonotes}
\usepackage[colorlinks=true, allcolors=blue]{hyperref}

\title{ARTH: Algorithm For Reading Text Handily - An AI Aid for People having Word Processing Issues}
\author[1]{Akanksha Malhotra}
\affil[1]{Harcourt Butler Technological Institute}
\author[2]{Sudhir Kamle}
\affil[2]{Indian Institute of Technology, Kanpur}
\date{}
\begin{document}
\maketitle

\begin{abstract}
The objective of this project is to solve one of the major problems faced by the people having word processing issues like trauma, or mild mental disability. “ARTH” is the short form of Algorithm for Reading Handily. ARTH is a self-learning set of algorithms that is an intelligent way of fulfilling the need for “reading and understanding the text effortlessly” which adjusts according to the needs of every user. The research project propagates in two steps. In the first step, the algorithm tries to identify the difficult words present in the text based on two features - the number of syllables and usage frequency - using a clustering algorithm. After the analysis of the clusters, the algorithm labels these clusters, according to their difficulty level. In the second step, the algorithm interacts with the user. It aims to test the user’s comprehensibility of the text and his/her vocabulary level by taking an automatically generated quiz. The algorithm identifies the clusters which are difficult for the user, based on the result of the analysis. The meaning of perceived difficult words is displayed next to them. The technology "ARTH" focuses on the revival of the joy of reading among those people, who have a poor vocabulary or any word processing issues.
\end{abstract}
\section{Introduction}

There are about 1 percent of people of the world suffer from intellectual disability. About 85 percent of people with intellectual disabilities fall into the mild category. A person who can read, but has difficulty comprehending what he or she reads represents one example of someone with mild intellectual disability. Similarly, about 10 percent of people with intellectual disabilities fall into the moderate category. They can understand simple sentences but have difficulty in comprehending complex structures. This constitutes about 72 million people in the world. Almost all of these people cannot continue their education after  3rd to 6th grade. Due to their poor attention span and intricacies of searching word meanings in dictionaries, they quit reading. ARTH aims to revive the joy of reading amongst such students.

The main agenda of the research is to identify those words with which user does not feel comfortable and requires help. 

\subsection{User Persona}
Misha is a mildly mentally disabled child of age 15 years. She is declared unfit to continue further schooling and left her school in 3rd grade. Her mother tried to teach her at home, but due to her poor vocabulary level, it took a lot of time to read one page of her textbook. Thus, she gave up on her reading habits even though she loved reading. This technology aims to revive the joy of reading/learning in the lives of such people. With this technique, she can read stuff on any reading device, as the meaning of the words which are unknown to her are displayed next to them while she is reading, thereby not hindering her experience.

\section{Algorithm}
ARTH works in two stages. Let them call it pre-user interactive stage and user-interactive stage.
Steps in the Pre-user interactive stages:
\begin{enumerate}
\item Text Preprocessing using natural language processing
\item Syllabification of Words
\item Usage frequency calculation
\item Using Clustering algorithm to measure difficulty levels
\end{enumerate}
Steps in the User-interactive stage:
\begin{enumerate}
\item Automated quiz generation from the text.
\item Quiz evaluation.
\item Generation of new text with meaning appended next to words deemed difficult for the user.
\end{enumerate}
See the algorithm in the Figure \ref{fig:1} 

\subsection{Text Preprocessing using Natural Language Processing}
The algorithm starts with the reading of the raw text file. The document received is preprocessed, to make it ready for further steps. It uses algorithms for natural language processing to solve the designated task. The process can be viewed graphically by using flowchart in the Figure \ref{fig:2} :

\begin{figure}
\centering
\includegraphics[width=0.8\textwidth]{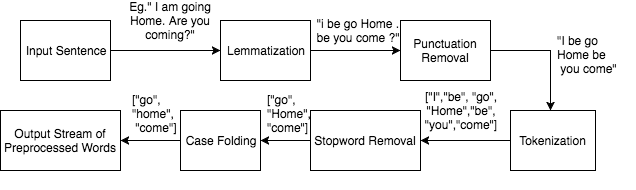}
\caption{\label{fig2:}   Text Preprocessing flowchart}
\end{figure}

Lemmatization and stemming are two algorithms that deal with the process of reduction of inflectional forms and sometimes derivationally related forms of a word to a common base form. Stemming refers to a crude heuristic process that chops off the ends of words in the hope of achieving this goal correctly most of the time and often includes the removal of derivational affixes. Lemmatization usually refers to doing things correctly with the use of vocabulary and morphological analysis of words, typically aiming to remove inflectional endings only and to return the base or dictionary form of a word, which is known as the lemma.

Tokenization is the task of chopping document up into words, also called tokens, perhaps at the same time throwing away certain characters, such as punctuation. A token is an instance of a sequence of characters which becomes a useful semantic unit for processing in some particular document. Tokenization in other terms is the detection of the end of the word and start of next word. Therefore, tokenization is sometimes also referred as word segmentation.

The stopwords are the words frequently occurring in most of the texts.  The general strategy for determining a stop word is to sort the terms by collection frequency (the total number of times each term appears in the document collection), and then to take the most frequent terms, often hand-filtered for their semantic content relative to the domain of the documents being indexed, as a stop list , the members of which are then discarded during indexing. In ARTH case,  a generalized list of stopwords mostly consisting of articles, prepositions, etc are used.

\subsection{Parameters for Detection of Difficulty Level of A Word}

ARTH uses two parameters for the detection of difficulty, 

Syllables: Syllables have a direct relation to the difficulty of the word. It is defined as a unit of pronunciation having one vowel sound, with or without surrounding consonants, forming the whole or a part of a word. Therefore an increase in the number of syllables will cause an increase in the difficulty of the word.

Usage Frequency: The usage frequency is indirectly related to the difficulty of the word. More is the usage of a particular word, more likely it is that the user will be familiar with the word.

\subsection*{Syllabification}

Syllabification is a method to know the number of syllables in a word. A syllable is a unit of organization for a sequence of speech sounds. A syllable tells us how troublesome it is for a person to enunciate a word. Greater the number of syllables, harder it is to pronounce and understand them. Thus, the number of syllables is chosen as a parameter to decide the difficulty of word. There are different ways of doing syllabification:

\begin{enumerate}
\item Rule Based Approach
\item Machine Learning Approach
\end{enumerate}

Using rule-based approach, we can actually develop a fine algorithm. I researched a lot about various rules that can tell about the number of syllable in a word[2]. The accuracy of above approach is 82.5\%

The calculation of the number of syllables in a word can be regarded as the classification problem where for a particular word we have to predict the number of syllables (class), which is essentially a set of discrete classes. For the experimentation purpose, I have used the most basic classifier of all the classifiers known, called Naive Bayes Classifier. 

\begin{table}
\centering
\begin{tabular}{l|r}
Method &Accuracy \\\hline
Rule Based Approach & 82.5\% \\
Naive Bayes Approach &  82.11\%
\end{tabular}
\caption{\label{tab:widgets}An example table.}
\end{table}

\subsection*{Calculation Of Usage Frequency}
The words, which we read occasionally are more familiar and easily understood by the people. By the term usage of the word, we mean the number of times a word has occurred in the texts available online. The relationship between usage and difficulty of the word is inverse. The masses quickly understand the frequently used words, but the meaning of less commonly used words has less probability of being known to the user. For calculation of usage, a database of the pool of words with their frequency is stored. All the words read in the text are searched in this database and their corresponding frequency is stored.

 \section{Clustering of Words Based on Number of Syllables and Usage Frequency}

\subsection*{Normalization of features}
Normalization is a re-scaling technique. The maximum value usage frequency feature can go up to millions, while the highest value of syllable feature is 12. There is a huge gap in the features. Since we are going to use these features for a clustering, so it is important to rescale them. As in K-means, we use Euclidean distance, and if we want that all the features to contribute equally, then we have to do normalization. 
There are two methods of rescaling the features:
\begin{enumerate}
\item  Min - Max scaling: In this score, the data is rescaled to lie within the fixed range, usually 0 to 1. The effect of outliers is diminished, when we use min-max scaling
					\[X_{norm}=\frac{X-X_{min}}{X_{max}-X_{min}}\]
\item Z-Score:The features will be rescaled so that they’ll have the properties of a standard normal distribution with $\mu = 0$ and $\sigma = 1$ , where $\mu$  is the mean (average) and  $\sigma$ is the standard deviation from the mean; standard scores (also called z-scores) of the samples are calculated as follows:	
\[z=\frac{X-\mu}{\sigma}\]
\end{enumerate}		

Z-score is usually a preferred method while dealing with the clustering algorithm and hence z-score is used in the project for normalization.

\subsection*{K-Means Clustering}
Unsupervised machine learning is the machine learning task of inferring a function to describe hidden structure from "unlabeled" data. We try to find structure in the dataset using unsupervised learning. There are many algorithms for unsupervised machine learning.  In ARTH for clustering of words into different levels of difficulty clustering algorithm is used. There are a lot of clustering algorithms like K-means, DBSCAN, Agglomerative Clustering, etc. I have implemented K-means clustering.

The K in K-means clustering is based on length of the text entered. If the length is greater than 400 words we keep K as 5 else we keep as 3.

\begin{figure}
\centering
\includegraphics[width=0.5\textwidth]{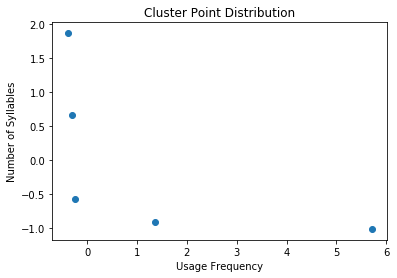}
\caption{\label{fig3:}   Cluster Point Distribution}
\end{figure}

After dividing words into K clusters, we then try to identify vocabulary level of users.

 \section{Automated Quiz Generation}
Automated Quiz will help in analyzing the vocabulary of the person. The aim of this application is to ask questions from the text the person is going to read. Each time there will be a different text, thus a predefined structure cannot be used. Hence, the need for the automated quiz generation. 
 The quiz generator mostly deals with:
\begin{itemize}
\item Question Formation.
\item Correct Answer Generation.
\item Wrong Options generator. 
\end{itemize}

\subsection*{Question Formation:} 
The question formation algorithm is a brute force algorithm due to limited time constraint. In future, will research and make it better. Algorithm can be divided into following steps:

\textbf{Step - 1 :  Words Selection}
We want to check the vocabulary level of the person so we choose the words. For this reason, we select randomly four words from the clusters obtained in previous step.

\textbf{Step 2 : Selection of sentences having those words}
The text is tokenized into sentences and select the sentence which have randomly chosen words.

\textbf{Step 3: Creation of questions}
We create a template for question generation, “What is the meaning of” + word + “ in the sentence” + sentence + “?”  

\subsection*{Correct Option Generator}
Correct option generator, generates answer for the question asked. We can use a dictionary based approach to find the correct meaning of the words. But there is a problem in this approach. Consider following two examples:
I am going to bank for depositing money.
I am going to bank for picnic.
In first sentence, the word “bank” means a financial establishment. In second sentence, the word “bank” means the land sloping down to a river or lake. A single word can be used in multiple ways, having multiple meaning.  This problem is known as Word Sense Disambiguation Problem in NLP. We implemented simplified Lesk algorithm for ARTH.

\begin{figure}
\centering
\includegraphics[width=0.8\textwidth]{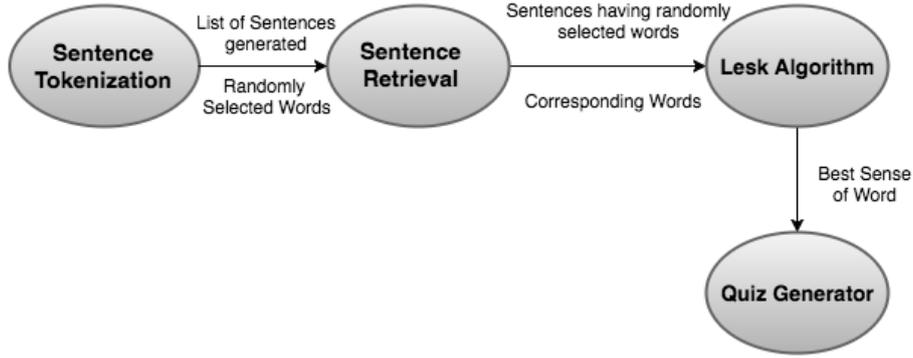}
\caption{\label{fig4:}   Correct Option Generator}
\end{figure}
\subsection*{Wrong Option Generator}
The task of wrong option generator is to generate wrong options so that the choices can be emerged. Steps involved in wrong option generator:
\begin{itemize}
\item Antonym Generation: Using Wordnet we can find the antonym of the words thus generating one wrong example.
\item Find those synonyms of words, using Synsets of words from Wordnet, whose meaning doesn’t matches to the word. For measuring the similarity scores, we have used similarity metrics.
\item If similarity  measures is greater than 5 then we accept it as wrong answer and add it to the list.
\item We randomly select words from the vocabulary and try to find similarity between them, if they are not similar, we include the word in our list.
\end{itemize}

Wu \& Palmer measure (wup) : The Wu \& Palmer measure (wup) calculates similarity by considering the depths of the two concepts in the UMLS, along with the depth of the LCS. The formula is 
\[score = \frac{2*depths(lcs)}{ depth(s1) + depth(s2)}\]
This means that 0 < score <= 1. The score can never be zero because the depth of the LCS is never zero (the depth of the root of a taxonomy is one). The score is one if the two input concepts are the same.
\begin{figure}
\centering
\includegraphics[width=0.8\textwidth]{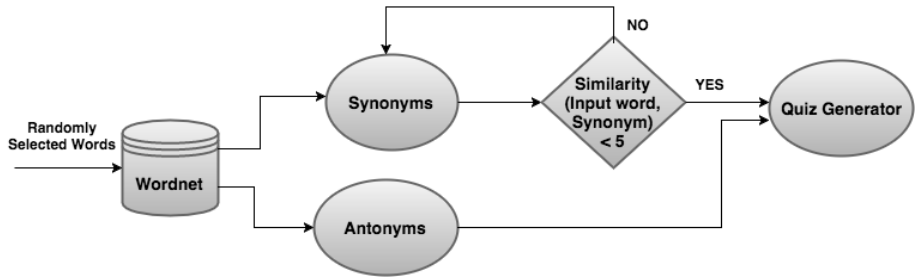}
\caption{\label{fig5:}   Wrong Option Generator}
\end{figure}
\subsection{Quiz Generation}
After the formation of questions, their correct answers and other wrong options, quiz can be generated in the UI for the user. The process of automated quiz generation is shown in the figure  {fig:6} :
\begin{figure}
\centering
\includegraphics[width=0.8\textwidth]{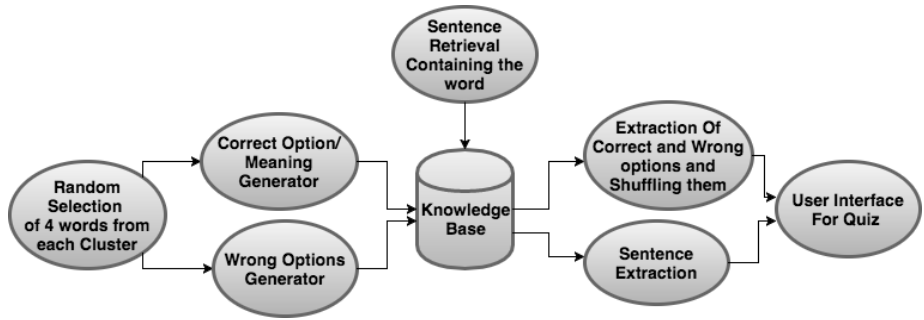}
\caption{\label{fig6:}   Quiz Generator}
\end{figure}
\subsection{Analysis of Quiz} 
The answers given by users are evaluated. As four questions represents one cluster. If for a particular cluster, the person answers correctly 75\% of the questions, then we mark the cluster easy for the user. For those clusters, the person gives all wrong answers or only answers 25\%  correctly, we declare them as hard for user. If a person answers 50\% correctly, then we again try to cluster that particular cluster into two groups, and ask quiz of six questions, if he answers 2 correct questions from a particular group, the group is labeled as easy for the person. 

\section{Generation of new text}
The new text can be generated by identifying the sentences that contains words from the clusters marked as hard. We can then use Lesk algorithm to find the best sense of the word. We can now generate a new text with the best sense or meaning written next to the difficult words inside a bracket. Now a user never has to refer a dictionary if he doesn’t know the word.

\section{Present: ARTH's Uniqueness and Difference from the Other Scales }

In the world, not everyone is blessed with a good health. Just because a person has a disability, we cannot revoke his /her right to education. If he/she cannot read in a normal way, we can make them revive joy of reading in a new way. This is what ARTH is trying to do. Where all the above scales focus on a generalized approach of finding the difficulty of the whole text and determining the age limit for which the text is suitable, ARTH focuses on making the text suitable for its user, thereby removing all barriers that will come in his/her way of reading. ARTH understand that each user is unique and has different needs. So rather than creating a generalized scale, it gauges the understanding level of its user and helps them to understand the text by showing the meaning of the sentences. 

\section{Further Research Areas}
In future, we can enhance ARTH by extending it to document level from word level, i.e., solving it as a Text simplification problem. Further,  We can go in depth is automated question answering mechanism, we have not dealt with the subject in more detail than I should have, due to time constraints. 

\section{References}

[1] Christopher D. Manning, Prabhakar Raghavan and Hinrich Schütze, The Introduction to Information Retrieval, $http://nlp.stanford.edu/IR-book/$

[2] Sebastian Raschka, (2014, October 4). Naive Bayes and Text Classification \- Introduction and Theory [Blog post]. Retrieved from $https://sebastianraschka.com/Articles/2014_naive_bayes_1.html$

[3] Brett Lantz( 2013) , Machine Learning with R, Birmingham: Packt Publishing Ltd 

[4] Trevor Hastie, Robert Tibshirani and Jerome Friedman (2008), The Elements of Statistical Learning, California: Springer

[5] Jacob Perkins,  Python Text Processing with NLTK 2.0 Cookbook,  Packt Publishing

[6] Martin Ester, Hans-Peter Kriegel, Jiirg Sander, Xiaowei X, 1996, A Density-Based Algorithm for Discovering Clusters in Large Spatial Databases with Noise, KDD-96 Proceedings

[7]  Pedregosa, F. and Varoquaux, G. and Gramfort, A. and Michel, V. and Thirion, B. and Grisel, O. and Blondel, M. and Prettenhofer, P. and Weiss, R. and Dubourg, V. and Vanderplas, J. and Passos, A. and Cournapeau, D. and Brucher, M. and Perrot, M. and Duchesnay, E., Scikit-learn: Machine Learning in Python, Pedregosa et al., JMLR 12, pp. 2825-2830, 2011.

[8] Dan Jurafsky and James H. Martin, Speech and language Processing. Retrieved From,
$https://web.stanford.edu/~jurafsky/slp3/slides/Chapter18.wsd.pdf$

[9] Rudell, Alan P. (1993, Dec 01). Frequency of word usage and perceived word difficulty: Ratings of Ku{\v{c}}era and Francis words. Behavior Research Methods, Instruments, {\&} Computers,  25, 4, 455-463. doi:10.3758/BF03204543

[10] Susanti, Yuni {\&} Iida, Ryu {\& } Tokunaga, Takenobu. (2015). Automatic Generation of English Vocabulary Tests. 1. 77-87. 10.5220/0005437200770087. 

\end{document}